%% file: main.tex
\documentclass[runningheads]{llncs}
\usepackage[T1]{fontenc}
\usepackage{graphicx}
\usepackage{hyperref}
\usepackage{latexsym}

\newcommand{\para}[1]{\paragraph{\textnormal{\textbf{#1}.}}} 

\newcommand{\knn}{$k$-NN}
\newcommand{\shade}{\cellcolor{lightgray}}
\newcommand{\rot}[1]{\rotatebox[origin=c]{90}{#1}}

\usepackage[titlenumbered,ruled,vlined]{algorithm2e}
\usepackage[normalem]{ulem}

\usepackage{tabularx}
\usepackage[export]{adjustbox}
\usepackage{subcaption}
\usepackage{booktabs}
\usepackage{multirow}
\usepackage{xcolor,colortbl}
\usepackage{float}
\usepackage{siunitx}

\usepackage{calrsfs}
\DeclareMathAlphabet{\pazocal}{OMS}{zplm}{m}{n}
\DeclareMathAlphabet{\pazobfcal}{OMS}{cmsy}{b}{n}
\newcommand{\vect}[1]{\mathbf{#1}}

\usepackage{amssymb}
\usepackage{amsmath}
\DeclareMathOperator*{\argmax}{arg\,max}

\usepackage{enumitem} 

\setitemize{noitemsep}
\setlist{topsep=0pt, leftmargin=*}

\newcommand{\uls}{\begin{itemize}[leftmargin=*]}
\newcommand{\ule}{\end{itemize}}
\newcommand{\ols}{\begin{enumerate}[leftmargin=*]}
\newcommand{\ole}{\end{enumerate}}
\newcommand{\li}{\item}

\newboolean{showcomments}
\setboolean{showcomments}{true} 
\ifthenelse{\boolean{showcomments}}
{
	\newcommand{\nb}[3]{
		{\colorbox{#2}{\bfseries\sffamily\scriptsize\textcolor{white}{#1}}}
		{\textcolor{#2}{$\blacktriangleright$\textsf\small{#3}$\blacktriangleleft$}}}
	 
}{
	\newcommand{\nb}[3]{}
} 

\begin{document}
\title{One size doesn't fit all: Predicting the Number of Examples for In-Context Learning}
\titlerunning{Adaptive In-Context Learning}

%
\author{Manish Chandra\inst{1}\orcidID{0009-0000-6156-5337} \and
Debasis Ganguly\inst{1}\orcidID{0000-0003-0050-7138} \and
Iadh Ounis\inst{1}\orcidID{0000-0003-4701-3223}}
\authorrunning{M. Chandra et al.}
%
\institute{University of Glasgow, Glasgow, United Kingdom
\email{m.chandra.1@research.gla.ac.uk, Debasis.Ganguly@glasgow.ac.uk, iadh.ounis@glasgow.ac.uk}\\
}

\maketitle              
\begin{abstract}

In-context learning (ICL) refers to the process of adding a small number of localized examples from a training set of labelled data to an LLM's prompt with an objective to effectively control the generative process seeking to improve the downstream task performance. Existing ICL approaches use an identical number of examples (a pre-configured hyper-parameter) for each data instance. Our work alleviates the limitations of this `one fits all' approach by dynamically predicting the number of examples for each data instance to be used in few-shot inference with LLMs. In particular, we employ a multi-label classifier, the parameters of which are fitted using a training set, where the label for each instance in this training set indicates if using a specific value of $k$ (number of most similar examples from 0 up to a maximum value) leads to correct $k$-shot downstream predictions.
Our experiments on a number of text classification benchmarks show that AICL substantially outperforms standard ICL by up to 17\%.

\end{abstract}
\section{Introduction}
\input{sections/introduction}

\section{Related Work}
\input{sections/related_work}

\section{Proposed Methodology} \label{sec:methodology}
\input{sections/methodology}

\section{Evaluation} \label{sec:expsetup}
\input{sections/experiment_setup}
\para{Concluding Remarks}
\input{sections/conclusion}

\begin{credits}
\subsubsection{\discintname}
The authors have no competing interests to declare that are relevant to the content of this article.
\end{credits}



%
%
%
\bibliographystyle{splncs04}
\bibliography{custom}
%




\end{document}

%% file: sections/introduction.tex
Large Language Models (LLMs) exhibit remarkable abilities to model text semantics in an abstract and general manner without any task specific training \cite{radford}. 
\textbf{In-context learning} (\textbf{ICL})\footnote{Also interchangeably known as few-shot learning or retrieval-augmented generation (RAG) with ground-truth labels.} makes use of this abstract representation and knowledge representation capabilities of LLMs \cite{gpt3,arora2022ask,weidinger2022taxonomy} to address a range of different downstream tasks. For instance, such models can provide effective solutions for tasks including assessing reviews \cite{mysore2023large}, answering questions \cite{li-etal-2023-shot}, recommending relevant documents \cite{pradeep2023does} etc., by using only a small number of (even zero) examples without any task-specific training.

\input{figdefs/example_workflow}

More formally, ICL refers to the process of conditioning an LLM's decoder (frozen parameters) towards generating potentially relevant outputs that could then be interpreted in the context of a specific task, e.g., words such as `great' and `wow' can be mapped to a positive sentiment \cite{schick-etal-2020-automatically}. The output of an LLM's decoder is controlled by varying the input to the LLM, which is usually structured in the form of an instruction (the task description), and a small number of representative examples \cite{ni-etal-2022-large}.
Figure~\ref{fig:icl-workflow} shows an example of how ICL works for the downstream task of movie review sentiment analysis. It can be seen that both demonstrations (shown in blue) and the current instance form a part of the input to an LLM.
%
%
%
Researchers have worked towards adapting an ICL workflow in several ways, which range from using localised examples \cite{liu-etal-2022-makes} instead of random ones, to diversifying these examples \cite{levy-etal-2023-diverse}, and also ideally ordering them \cite{icl-reordering,rubin-etal-2022-learning}.
%
%
What prior work has not addressed so far is the effect of \textbf{dynamically selecting the number of examples} in an ICL workflow.
Our idea takes motivation from information retrieval (IR), where different queries exhibit different levels of retrieval performance mainly due to the inherent characteristic of the information need itself \cite{Kanoulas:2011}, or how well-formulated the query is \cite{deep_qpp}. 
Drawing the parallel that a query in IR is analogous to a test instance in ICL \cite{icl_perspective} and that the localised examples are potentially relevant documents \cite{rubin-etal-2022-learning}, we hypothesise that some test instances are associated with better candidates for training examples (i.e., examples which are useful in the sense that including them as a part of the prompt leads to correct predictions), and hence including a small number of these examples should be adequate. On the other hand, the retrieval quality for some test instances used as queries do not yield good candidates, as a result of which, one needs to further look down the ranked list to collect the useful ones \cite{choppy,vdp}.
The parallel with IR means that the notion of relevance of a document to a query needs to be replaced by the downstream usefulness of an example to a test instance.

The idea of choosing a variable number of localised examples is also somewhat similar to selecting a variable-sized neighborhood for \knn~classification \cite{10.1145/3055635.3056604}, where the key idea is that a homogeneous neighborhood is likely to require a relatively small-sized neighborhood for a correct prediction, and a heterogeneous one would likely require a larger one (see Figure~\ref{fig:k_in_knn}). The same idea can be applied to ICL, where a test instance that is similar to a number of training instances with conflicting labels may require a larger number of examples.


In our work, we apply a supervised learning based workflow to learn a mapping between the features of a test instance (embedding plus the label distribution of its neighborhood) and the ideal number of examples that should be used in ICL for the downstream prediction. More specifically, for each instance of the training set, we vary $k$ - the number of ICL examples - within a range of $0$ to a pre-configured maximum value (say $M$) and store each indicator of whether a particular value of $k$ leads to a correct prediction as an $(M+1)$-length Boolean vector (e.g., the indicator for $k=1$ in Figure \ref{fig:icl-workflow} is negative, whereas the one for $k=2$ is positive). We then train a multi-label classifier on these pairs of instances and the Boolean indicators, the assumption being that similar instances with similar label distributions should also exhibit a similar distribution over the values of $k$ leading to correct $k$-shot learning. During inference time, for each test instance, we apply the multi-label classifier to obtain a candidate list of predicted $k$ values, and we select the one for which the prediction confidence is the highest. We then use these many examples for the ICL-based downstream prediction.
This means that as per the schematics of Figure \ref{fig:icl-workflow}, we can potentially find out a `green path' leading to a correct prediction for each test instance among several other `red paths' that lead to incorrect ones.

%% file: figdefs/example_workflow.tex
\begin{figure}[t]
\centering
\begin{subfigure}[b]{.64\columnwidth}
\includegraphics[width=.99\columnwidth]{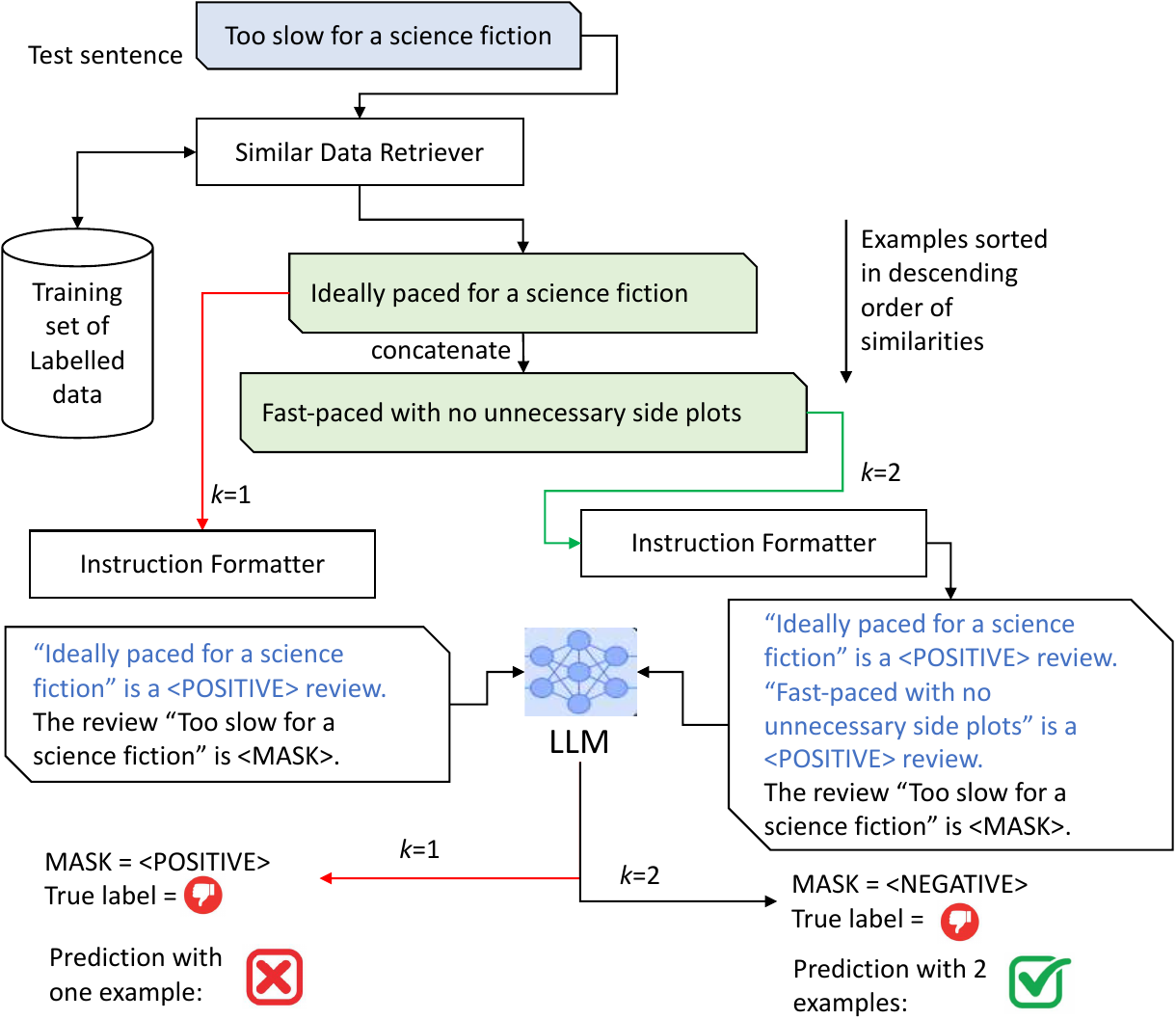}
\caption{
\label{fig:icl-workflow}
}
\end{subfigure}
\begin{subfigure}[b]{0.26\columnwidth}
\includegraphics[width=0.99\columnwidth]{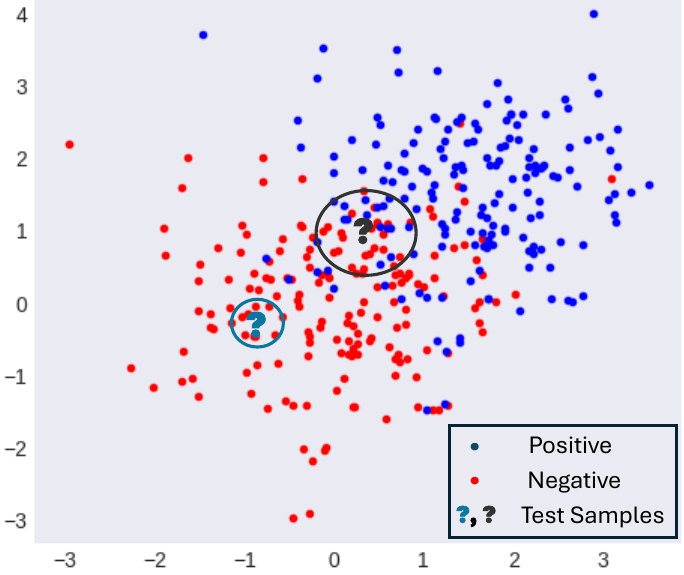}
\caption{
\label{fig:k_in_knn}
}
\end{subfigure}
\caption{
\textbf{a)} \textit{Example workflow of ICL for sentiment classification}: The example shows a test instance for which a single demonstration (as retrieved from the training set) does not result in a correct prediction (prediction workflow of the red arrows). It also shows that increasing the number of demonstrations from one to two results in a correct prediction (green arrows). We propose a method to estimate the number of examples that are likely to yield a correct prediction.
\textbf{b)}
\textit{Motivation behind using a variable number of examples for ICL across the test instances}: The test sample `\textcolor{teal}{\textbf{?}}' is located within a homogeneous neighborhood of negative data points indicating that an LLM may perform well with only a few nearest neighbour demonstrations. The test instance `\textcolor{darkgray}{\textbf{?}}', on the other hand, is located within a heterogeneous neighborhood, as a result of which, an LLM may require a higher number of such demonstrations to correctly predict its class. 
}
\end{figure}

%% file: sections/related_work.tex
\para{Prompt tuning and searching}
LLMs, when scaled from millions to billions of parameters, have been demonstrated to be adaptable to a broad set of tasks due to instruction tuning \cite{NEURIPS2022_b1efde53,gpt3}, in the sense that they are not only able to produce semantically correct and coherent text, but are also able to adapt themselves surprisingly well with small changes in contexts supplied as inputs, commonly called prompts \cite{arora2022ask}.
Previous research studies the problem of constructing an appropriate prompt for LLMs from two broad perspectives: a) \textit{prompt tuning} in the
embedding space \cite{li-liang-2021-prefix,liu-etal-2022-p,qin-eisner-2021-learning,LIU2023}, and b) \textit{prompt searching} in the text space \cite{lu-etal-2022-fantastically,zhang2023automatic,diao2023active,liu-etal-2022-makes,llm_distracted}.
Prompt tuning is a lightweight alternative to fine-tuning, which keeps language model parameters frozen and instead optimizes a sequence of continuous task-specific vectors. The key idea of prompt tuning is to inject task-specific embedding into hidden layers and then tune these embeddings using gradient-based optimization. However, these methods require the modification of the original inference process of the model, which is impractical for the case of black-box LLM services, such as GPT3 \cite{gpt3}. Furthermore, prompt tuning introduces additional computational and storage costs, which is typically expensive for LLMs. A more efficient way is to optimize prompting via searching appropriate demonstration samples and ordering them in the original text space.

\para{In-context Learning (ICL)}
ICL has advanced the use of LLMs for task-specific applications with minimal examples, forming a solid foundation for subsequent investigations into optimizing this learning paradigm \cite{dong2022survey}. 
In ICL, a small number of labeled examples from a training set are appended to a prompt instruction to control the text generation of LLM so that it is beneficial to a downstream task \cite{mysore2023large,li-etal-2022-encoder,ni2021large,pradeep2023does}. In addition to leveraging ICL for a purely generative task, e.g., question answering or abstractive summarisation \cite{gpt3,li-etal-2023-shot,tang-etal-2023-context}, a more common use is in a predictive task, such as text classification \cite{lu-etal-2022-fantastically,milios-etal-2023-context,wei2021few}, where each class is specified by a set of words, commonly called a verbaliser \cite{schick-schutze-2021-exploiting}.
Once each class for a predictive task is well-defined, the generated text can be mapped to the most likely class by using the probabilities of the tokens generated by the decoder.

A crucial aspect of ICL is the adept selection and utilization of the demonstrations for task comprehension. Recent works have explored the dynamics of in-context examples, elucidating how ICL enhances the language model's performance across various NLP tasks by providing a minimal set of examples at inference time \cite{han2023understanding}.
Furthermore, the effectiveness of ICL is significantly shaped by the strategies for selecting and ordering demonstrations. For instance, the work in \cite{luo2023dr} expanded the applicability of retrieval-based ICL approaches by demonstrating that even simple word-overlap similarity measures such as BM25 outperform randomly selected demonstrations.
Methods such as KATE \cite{liu-etal-2022-makes} also showed that localised examples work better than random ones.
Different from KATE that uses a fixed-size neighbourhood for each data instance, we propose to modify the standard ICL workflow by employing a variable number of examples.


%% file: sections/methodology.tex

\input{figdefs/arch}

\subsection{Standard In-Context Learning (ICL)}


In-context learning (ICL), unlike supervised learning, does not involve training a set of parameters on labeled examples. Rather, the posteriors are now a function of the following: a) text of the input test instance, b) the decoder parameters of a pre-trained LLM, c) a prompt instruction, and d) optionally, a set of $k$ input examples, commonly called $k$-shot prompting \cite{liu-etal-2022-makes}. Formally,
\begin{equation}
P(y|\vect{x}, k) = f(\vect{x}, \pazocal{N}_k(\vect{x}); \phi_{\text{LLM}}),
\label{eq:icl}
\end{equation}
where, different from a supervised setup, the function $f$ does not have a parameterized representation that can be learned using a training set with gradient descent. The function itself depends on the pre-trained frozen parameters $\phi_{\text{LLM}}$ of an LLM, the current inputs for which a label is to be predicted, and a prompt comprising a set of $k$ text units denoted by $\pazocal{N}_k(\vect{x})$.
%
%
This set $\pazocal{N}_k(\vect{x}) = \{\vect{z}_1, \ldots, \vect{z}_k\}$ of Equation~\ref{eq:icl} (where each $\vect{z}_j \in \pazocal{T}$, a training set of labelled data instances) in ICL is usually comprised of
localised examples, i.e., examples that are topically similar to the current instance \cite{liu-etal-2022-makes,luo2024incontext}.
%
Particularly for our experiments, we employ SBERT \cite{sbert} as the neighborhood similarity computation function as per the findings of \cite{liu-etal-2022-makes}.




\subsection{Adaptive ICL} \label{subsec:aicl}


We now describe our methodology that uses a variable number of examples by extending the standard ICL workflow of Equation \ref{eq:icl}. We call our method
`Adaptive In-Context Learning', or \textbf{AICL} in short.
The idea of AICL centres around choosing the context $\pazocal{N}_k(\vect{x})$ in a data-driven manner, i.e., making $k$ a function of the data, i.e., the current instance $\vect{x}$ itself. This is analogous to choosing a different value of $k$ for a $k$-NN based non-parametric model or choosing a different rank cut-off for top-retrieved set of documents for different queries \cite{choppy,vdp}. The motivation is that classifying some instances would be more difficult than others, in which cases they are potentially to benefit from a larger value of $k$ (more context). On the other hand, for relatively easier data instances, using too much context may be detrimental for an effective prediction.

Formally speaking, the difference of AICL with that of ICL (Equation~(\ref{eq:icl})) is that the value $k$, indicating the size of the neighborhood, is no longer a constant. Instead, we denote it by a parameterised function $\kappa(\vect{x})$ such that
\begin{equation}
P(y|\vect{x}, \kappa) = f(\vect{x}, \pazocal{N}_{\kappa(\vect{x})}(\vect{x}); \phi_{\text{LLM}}),
\label{eq:aicl}
\end{equation}
where $\kappa: \vect{x} \mapsto \{0,\ldots,M\}$, and $M$ is an upper bound on the number of example instances.


Figure~\ref{fig:arch} presents an overarching view of the AICL workflow.
In contrast to ICL, AICL involves an additional phase of training a classifier, $\kappa(\vect{x})$ of Equation~(\ref{eq:aicl}), to predict an appropriate number of examples by leveraging the training instances (the `classifier training' block in Figure \ref{fig:arch}). The `LLM inference' block shows the inference phase, where, given a test instance, we first predict the number of examples to be used and then follow the standard ICL procedure with those many examples fed as part of the additional context to the LLM.




\para{Obtaining the ground-truth values of the number of ICL examples for each training set instance}
For each training set instance $\vect{x} \in \pazocal{T}$, we first execute $k$-shot inference with progressively increasing values of $k$ within $\{0,\ldots,M\}$, where $M$ is a pre-configured threshold (specifically, $M=10$ in our experiments). This is depicted by the progressively increasing colored dotted regions within the `classifier training' block of Figure \ref{fig:arch}.
After obtaining $M+1$ posteriors $P(\hat{y}|\vect{x})$ for an instance $\vect{x}$ of the training set, 
we construct an $(M+1)$ dimensional Boolean vector, the $i^{th}$ component of which indicates if using an $i$ number of ICL examples lead to a correct downstream prediction for the current instance $\vect{x}$ (note that since $\vect{x}$ is a training set instance, the knowledge of its true label $y(\vect{x})$ is available).
Formally speaking,
\begin{equation}
\pazocal{K}_i(\vect{x}) = \mathbb{I}[y(\vect{x}) = \argmax P(y|\vect{x}, i, \phi_{\text{LLM}})],
\label{eq:candidate}    
\end{equation}
where $\pazocal{K}(\vect{x})=\{\pazocal{K}_0(\vect{x}),\ldots,\pazocal{K}_M(\vect{x})\}$ is an $(M+1)$ dimensional Boolean vector (note that we allow provision for the number of examples to be 0 as well, thus allowing provision for zero-shot prediction).

\para{Training a multi-label classifier}

As a next step, we learn a parameterized map from the inputs to the $M+1$ dimensional indicators on the number of ICL examples, i.e.,
\begin{equation}
\theta: \vect{x} \mapsto \pazocal{K}(\vect{x}),
\label{eq:mlc}
\end{equation}
where $\pazocal{K}(\vect{x}) \in \{0,1\}^{M+1}$ as defined in Equation \ref{eq:candidate} and $\theta$ is a set of parameters learned via cross-entropy loss.
Our decision to use a multi-label classification approach is motivated from a set of initial experiments, where we found that it outperformed a multi-class classifier approach (which we thus do not report in the paper).
The most likely reason why a multi-label classifier works more effectively for this predictive task is that it allows provision to select the most likely candidate number of ICL examples from a number of alternatives rather than relying on a single predicted value of $\theta(\vect{x})$.

Of course, this also means that there needs to be a precise selection criteria to choose the number of examples to be used in the inference time from a list of choices, i.e., the result of the multi-label prediction where the predicted label is 1 (sigmoid posterior is higher than 0.5). We tried two different heuristics for this selection - i) choosing the smallest index from the $M+1$ dimensional vector with a predicted value of 1 (sigmoid > 0.5), and ii) selecting the component with the highest confidence (sigmoid posterior).
Again after conducting a set of initial experiments, we found that the second heuristics almost always outperforms the first one, and hence our workflow of Figure \ref{fig:arch} employs the max-confidence heuristic for predicting the number of ICL examples to be used during inference (see Equation \ref{eq:aicl}).
More formally,
\begin{equation}
\kappa(\vect{x}) = \argmax \theta(\vect{x}),    
\end{equation}
where $\theta(\vect{x}) \in [0,1]^{M+1}$ denotes the posteriors (sigmoid values) obtained from a multi-label classifier trained with the labels as defined in Equation \ref{eq:candidate}. 

\para{Distribution of the downstream-task class labels as additional features}
The multi-label classifier $\theta: \vect{x} \mapsto \pazocal{K}(\vect{x})$ means that the number of ICL examples depend only on the document content, or in other words, topically similar content potentially requires a similar number of ICL examples for accurate predictions. However, this does not take into account the class label distribution, i.e., the $y(\vect{x})$ values of the examples. In fact, it is reasonable to assume that an instance with a more homogeneous neighborhood (in terms of the class label distribution) likely indicates an easier instance and hence may require a smaller number of examples, whereas a more heterogeneous neighborhood may indicate a larger number of examples (e.g., see Figure \ref{fig:k_in_knn}). 

This hypothesis means that the
use of neighborhood class distribution as additional features to encode each training data instance ($\vect{x} \in \pazocal{T}$) is likely to lead to a better predictor $\theta$, which in turn, is likely to lead to better estimations of $\kappa(\vect{x})$ for AICL.
%
%
%
Specifically, to incorporate the class prior information to aid predicting the number of ICL examples, we append an $M$ dimensional vector $\vec{l} \in \mathbb{Z}_p^M$ ($p$ is the number of class labels) to each SBERT embedded input vector $\vect{x} \in \pazocal{T}$. The $i^{th}$ component of this vector contains the class label of the $i^{th}$ neighbor in $\pazocal{N}_M(\vect{x})$, i.e., $\vec{l}=\{y(\vect{z}_1),\ldots,y(\vect{z}_M)\}$.
With this modification, the multi-label classifier now learns: 
\begin{equation}
\theta: \vect{x} \oplus \vec{l} \mapsto \pazocal{K}(\vect{x}),
\label{eq:mlc-l}
\end{equation}
where $\vec{l}$ denotes the class label distribution of $M$ nearest neighbors, and $\oplus$ denotes the concatenation operator.

%% file: figdefs/arch.tex
\begin{figure*}[t]
\centering
\includegraphics[width=0.99\textwidth]{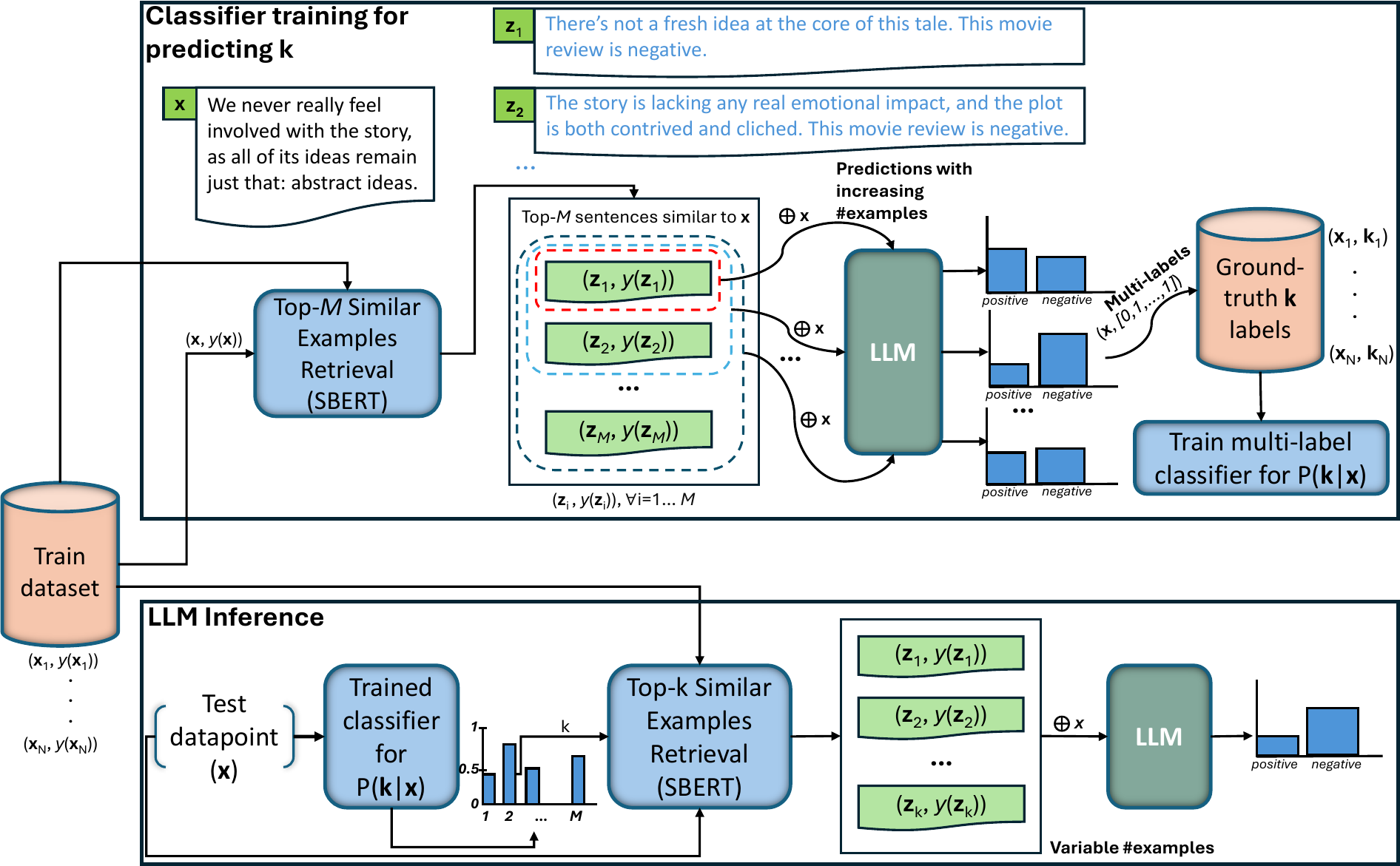}
\caption{
Schematic diagram of Adaptive In-Context Learning (AICL) workflow.}
\label{fig:arch}
\end{figure*}

%% file: sections/experiment_setup.tex

\input{tabledefs/datasets}
\subsection{Research Questions and Datasets} \label{subsec:dataset}

In Section \ref{sec:methodology}, we proposed a classifier-based approach to learn the optimal number of examples. In our experiments, we compare this approach for adaptive $k$-shot with fixed $k$-shot on standard datasets for text classification.
In particular, we investigate the following research questions.
\uls
\li 
\textbf{RQ-1}: Does adaptively selecting the number of examples in ICL (AICL) lead to improved downstream effectiveness?
\li 
\textbf{RQ-2}: Do the class labels of the neighbouring training set examples act as useful features to improve AICL performance? 
\li
\textbf{RQ-3}: Does AICL generalise well across different LLMs and datasets?
\li
\textbf{RQ-4}: Can the multi-label classification model (Equation \ref{eq:mlc-l}) be trained effectively on a small subsample of the training set, thus reducing the number of LLM calls during the training?
\ule

As per the setup of \cite{ma2023fairness,zhang-etal-2022-active}, we conducted our experiments on four standard text classification datasets, namely SST2, TREC \cite{trec}, CoLA and RTE. Except TREC, the other three datasets are part of the GLUE benchmark \cite{glue} 
exhibiting diversity in: i) domain, e.g., from movie reviews to news, ii) number of classes, ranging from 2 to 6, iii) average length of the documents ranging from sentences (e.g., 6.7 words on an average in CoLA) to relatively large text (e.g., 52.4 words on an average in RTE). This diversity in the characteristics of the respective datasets (see Table \ref{tab:task-instructions}) thus allows provision to assess the generalizability of our proposed approach. More details on each dataset is as follows.

\uls
\li \textbf{SST2}:
The Stanford Sentiment Treebank (SST) 
corpus consists of $11855$ sentences extracted from movie reviews. 
The SST2 (also called SST-binary) dataset is a subset of SST, specifically prepared for the task of binary sentiment classification by discarding neutral sentences. 

\li \textbf{TREC}:
The task is that of open-domain question classification of fact-based questions categorized into a total of 6 types \cite{ma2023fairness}, constituting $5500$ labeled questions in the training set and $500$ in the test set.

\li \textbf{CoLA}:
The Corpus of Linguistic Acceptability (CoLA) dataset consists of $10657$ sentences from $23$ linguistics publications annotated for grammatical correctness. It has a wide coverage of syntactic and semantic phenomena.

\li \textbf{RTE}:
The Recognizing Textual Entailment (RTE) dataset comes from a series of annual textual entailment challenges. The authors of the GLUE benchmark \cite{glue} combined the data from RTE1, RTE2, RTE3 and RTE5.
\ule



\subsection{Methods Investigated}

\para{Baselines}
We compare our proposed approach\footnote{Code available at \url{https://github.com/ManishChandra12/adaptiveICL}} with the following baselines:
\uls
\li \textbf{0-shot}: This approach tests a model’s innate capabilities, i.e., the capability to operate without any training data (information on class labels). Specifically, this requires setting $k=0$ in Equation \ref{eq:icl}.
\li \textbf{FICL} \cite{liu-etal-2022-makes} (Fixed ICL): This refers to the standard method of applying a static (fixed) number of localized (semantically similar) examples as input.
The number of examples to use in FICL is determined by optimizing $k \in \{1,\ldots,10\}$ by means of grid search on a validation set, which comprises $10\%$ of the training set (the same training set is also used in AICL).

\ule

\para{Variants of AICL}
We employ the following variants for our proposed methodology of employing a variable number of ICL examples. 
\uls
\li 
\textbf{AICL with Embedding-only - AICL(E)}: This is an ablation study, where the number of examples for each data instance is only a function of its textual content - more specifically its SBERT \cite{sbert} embedding (see Equation \ref{eq:mlc}).
\li 
\textbf{AICL with Embedding + Neighborhood class labels - AICL(E+N)}:
Similar to AICL(E) except that
the input to the multi-label classifier is now the embedding of a training data instance along with the class labels of its neighbors (see Equation \ref{eq:mlc-l}).
\ule

We also analyze the performance of AICL under an ideal setting, i.e., when one can choose the best prediction from among a set of predictions obtained with different values of $k$ in $k$-shot LLM-based inference by making use of the test set labels. Despite being non-pragmatic, this method provides an upper bound on the effectiveness that could, in principle, be obtained with the AICL framework. To avoid unfair comparisons of AICL* with AICL, we gray out the AICL* results indicating that these results are only to be used as an apex-line.

Corresponding to \textbf{RQ-4}, since constructing the ground-truth via Equation~\ref{eq:candidate} requires a total of $(M+1)$ LLM calls for each training instance, for efficiency reasons, we first train the multi-label classifier of the AICL workflow only on a subset of training data obtained by randomly sampling $50\%$ of data from each class.
Later on, as post-hoc analysis towards answering \textbf{RQ-4}, we also investigate if AICL results further improve (or degrade) with larger (or smaller) subsets sampled from the training data.

\subsection{Model and hyper-parameter settings}

To learn the number of examples to be used during inference, the AICL workflow relies on training the $\kappa(\vect{x})$ predictor (multi-label classifier of Equations \ref{eq:mlc} and \ref{eq:mlc-l}), which in turn, depends on the ground-truth labels ($\pazocal{K}$ of Equation \ref{eq:candidate}) of the number of examples that lead to correct predictions via LLM-based few-shot inference. But this means that these ground-truth labels depend on a particular choice of the LLM that is used during the training phase to check the accuracy of the downstream task labels (see Equation \ref{eq:candidate}).

To analyze the variations that may be caused due to this choice of LLMs, we conduct our experiments on three different LLMs - two each from the same family with different sizes, and one from a different family. The objective is to analyze the effects of variations in AICL's performance corresponding to the size of a model from the same family, or variations in characteristics in models across different families.
In particular, as instances of LLMs from the same family we use
Meta's Llama-2-7b and Llama-2-13b \cite{touvron2023llama2}, whereas as another LLM from a different family, we use Microsoft's Phi-2 \cite{phi}. Details on these models are as follows:
%
%
Llama-2 is a collection of open-source pretrained and fine-tuned text models
with a maximum context length of 4096 tokens (we use the 7B and the 13B parameter models).
Phi-2 \cite{phi} is a transformer with 2.7 billion parameters. It is trained on documents comprising Python code, StackOverflow, websites and natural language text generated by GPT-3.5.
Phi-2 with less than 3B parameters has been shown to yield comparable performance with those achieved with much larger number of parameter, e.g., the 13B variant of Llama-2. The maximum input length of Phi-2 is 2048 tokens. 

In our experiments, we set $M$ (the maximum number of ICL examples) to $10$ (see Equation \ref{eq:candidate}). Furthermore, for training the multi-label classifier parameters $\kappa(\vect{x})$ of Equations \ref{eq:mlc} and \ref{eq:mlc-l}, we choose a single hidden layer network with 64 neurons. These decision choices are based on initial experiments of hyper-parameter tuning conducted on the SST2 dataset with the Llama-2-7b model.

\input{tabledefs/results_long}

\subsection{Results}

\para{Main observations}

Table~\ref{table:results} shows a comparison between different ICL approaches on the four datasets.
The reported precision, recall and F-scores are averaged over 5 runs of the experiments. The number of examples in FICL, as shown in the corresponding column, were obtained by a grid search over $k$ from 1 to 10 on a validation set.
From Table~\ref{table:results}, 
it can be seen that AICL(E+N) turns out to be the best among the competing approaches - this method being statistically distinguishable than the FICL results (McNemar's test with 95\% confidence interval).
The likely reason it outperforms the baselines is that AICL is able to effectively adapt the number of examples to use, thereby preventing itself from the degradation effects of non-relevant (not useful) examples. In effect, it learns a latent relationship between the topical content and the quantity of context required to guide the decoder's output towards improving a downstream task. Our observations reveal that \textbf{RQ-1} is answered in the affirmative, i.e., an adaptive selection of the number of examples in ICL does improve the downstream effectiveness.

For the AICL approaches, we report the average number of examples, as predicted by the output of the multi-label classifier (Equations \ref{eq:mlc} and \ref{eq:mlc-l}), in the column named $k$. In contrast to the grid-searched integer $k$ values in FICL, for AICL the number of examples ($k$) is an average over the test instances, and hence is a non-integer.

One interesting observation is that the oracle version of AICL (AICL*) performs substantially better than AICL with much smaller values of $k$, which testifies the fact that selecting the correct value of $k$ independently for each test instance can lead to large improvements in few-shot prediction effectiveness. This also means that there is a scope to further improve the AICL workflow.

\para{Sensitivity Analysis of FICL (Fixed-$k$ ICL)}

Figure~\ref{fig:results} shows the F1-scores of FICL on the test splits of the different datasets with a varying number of demonstrations. Firstly, it can be observed from Figure \ref{fig:results} that there is no consistent improvement in the performance of FICL with an increase in $k$. Since the FICL results are sensitive to $k$, it is difficult to choose a particular value of $k$ for a given downstream task because such a grid search is not a practical solution as the ground-truth is not supposed to be known apriori. We already showed in Table \ref{table:results} that optimizing $k$ by grid search on a validation set produces worse results than AICL. Additionally, Figure \ref{fig:results} shows that the AICL results also outperform FICL with $k$ optimized on the test set. This indicates that AICL can, in principle, be deployed on any downstream task without requiring any hyper-parameter optimization of $k$.



\input{figdefs/results}

\para{Effect of neighbourhood}
Among the two variants of AICL, Table \ref{table:results} shows that the use of neighbor class labels (the `E+N' variant) yields better results than the ablated (`E' variant), which means that \textbf{RQ-2} is answered in affirmative.
In relation to \textbf{RQ-3}, Table \ref{table:results} shows that the improvements in AICL with respect to FICL is consistent across different datasets and also across different LLMs.


\input{figdefs/scalability}

\para{Multi-label classifier in AICL with reduced training data}
It is evident from Table~\ref{table:results} that for each dataset, the best value of $k$ in FICL differs from one LLM to another. This means to construct the correct ground-truth for training the multi-label classifier in AICL (Equation \ref{eq:mlc-l}), $M+1$ LLM predictions need to be made for each instance, which means that the total number of LLM invocations is $|\pazocal{T}|(M+1)$, where $\pazocal{T}$ denotes the training set.

A solution to reduce training time is to use a small subsample of the training data to learn the parameters of Equation \ref{eq:mlc-l}.
Figure~\ref{fig:scalability} shows the variations in F1 scores for different proportions of the training set used to learn $\kappa(\vect{x})$. While instance selection based sampling methods have been proposed in the literature \cite{instance_selection_survey}, for this work we simply use uniform sampling and report the results averaged over 5 runs of the experiments.

An interesting observation is that AICL is able to outperform the baseline with as small a proportion as 30\% (or even less in some cases) of the training set being used to train $\kappa(\vect{x})$, which means that \textbf{RQ-4} is answered in the affirmative. 

\para{Sensitivity to Labels}
We now analyze how AICL behaves in the absence of any labelled training data \cite{rethinking_role} - a setup commonly known as retrieval-augmented generation (RAG) \cite{ragReview}.
A way to simulate this configuration in our experiment setup is to keep only the content and exclude the labels from the demonstrations added to the prompt.
Equation~\ref{eq:candidate}, in its current form, cannot be used to get all the numbers of examples that work, because of the unavailability of labels. We therefore modify Equation~\ref{eq:candidate} to set $\pazocal{K}_i(\vect{x}) = 1$, where
$0 \leq i \leq M$ denotes the number of examples that lead to the maximum mutual information of the Softmax class posteriors, as obtained with $i$-shot LLM output. The rest of the components of $\pazocal{K}_i(\vect{x}) $ are set to 0, which
means that instead of a multi-label classification setting, for the unlabeled setup the $k$ predictor is now a multi-class classifier.
Also note that due to the unavailability of class labels, the AICL(E+N) version (Equation~\ref{eq:mlc-l}) cannot be executed in this RAG setup.
%
%


\input{tabledefs/results_nolabel}

The results of these additional experiments, carried out on the Phi-2 model, are shown in Table~\ref{table:results_nolabel}. From Table~\ref{table:results_nolabel}, it can be seen that absence of labels hurt performance - however, not by large margins, which is consistent with the findings of \cite{rethinking_role}.
%
AICL still continues to mostly be the best performing method even without the presence of any class labels. The only dataset where AICL does not yield the best performance is RTE, where the best performance is achieved with 0-shot indicating that labels are crucial for this dataset, and hence all RAG approaches fail on this particular dataset. It is encouraging to see, however, that AICL* (the oracle setting of AICL) without labels still continues to be substantially better than FICL (ICL with fixed context sizes).

%% file: tabledefs/datasets.tex
\begin{table}[t]
\centering
\small
\begin{adjustbox}{width=0.99\columnwidth}    
\begin{tabular}{llclc}
\toprule
Dataset & Task & \#Classes & Document Type & Avg Len (\#words)\\
\midrule
SST2 & Sentiment Analysis & 2 & Movie Reviews & 19.3\\
TREC & Question classification & 6 & Open Domain Questions & 10.2\\
CoLA & Grammatical Error Detection & 2 & News and Wikipedia & 6.7\\
RTE & Language Inference & 2 & Miscellaneous & 52.4\\
\bottomrule
\end{tabular}
\end{adjustbox}
\caption{The characteristics of the datasets used in our experiments.}
\label{tab:task-instructions}
\end{table}

%% file: tabledefs/results_long.tex
\begin{table}[t]
\centering
\begin{adjustbox}{width=0.95\textwidth}
\begin{tabular}{@{}lcS[table-format=2.2]lllS[table-format=2.2]lllS[table-format=1.2]llr@{}}
\toprule
\multirow{2}{*}{} &  & \multicolumn{4}{c}{Llama-2-7b} & \multicolumn{4}{c}{Phi-2} & \multicolumn{4}{c}{Llama-2-13b} \\
\cmidrule(r){3-6} \cmidrule(r){7-10} \cmidrule(r){11-14}
& Method & $k$ & Prec & Rec & F1 & $k$ & Prec & Rec & F1 & $k$ & Prec & Rec & F1 \\
\midrule
\multirow{5}{*}{\rot{SST2}} & 0-shot & 0 & .7227 & .6123 & .5589 & 0 & .8988 & .8919 & .8914 & 0 & .7341 & .6443 & .5891 \\  		
 & FICL & 9 & .8744 & .8687 & .8682 & 10 & .9279 & .9253 & .9252 & 9 & .8912 & .8733 & .8809 \\ 
 & AICL(E) & 8.63 & .9089 & .8946 & .8925 & 6.39 & .9307 & .9300 & .9300 & 5.73 & .9175 & .9065 & .9028 \\ 
 & AICL(E+N) & 8.23 & \textbf{.9099} & \textbf{.8964} & \textbf{.8954} & 5.64 & \textbf{.9350} & \textbf{.9345} & \textbf{.9345} & 4.89 & \textbf{.9189} & \textbf{.9071} & \textbf{.9034} \\ 		
 & \shade AICL* & \shade 1.27 & \shade .9765 & \shade .9753 & \shade .9753 & \shade .17 & \shade .9865 & \shade .9863 & \shade .9863 & \shade .84 & \shade .9866 & \shade .9859 & \shade .9848 \\ 		
\cmidrule{2-14}
\multirow{5}{*}{\rot{TREC}} & 0-shot & 0 & .1697 & .1687 & .0100 & 0 & .6289 & .4273 & .3526 & 0 & .3212 & .3685 & .3016 \\ 
 & FICL & 9 & .7388 & .7324 & .6608 & 8 & .6842 & .6582 & .6192 & 9 & .7612 & .7922 & .7071 \\ 				
 & AICL(E) & 9.60 & .7654 & .8049 & .7325 & 4.72 & .7146 & .7549 & .7196 & 5.51 & .7659 & .8218 & .7383 \\ 
 & AICL(E+N) & 9.58 & \textbf{.7682} & \textbf{.8075} & \textbf{.7357} & 5.57 & \textbf{.7254} & \textbf{.7673} & \textbf{.7291} & 4.93 & \textbf{.7737} & \textbf{.8288} & \textbf{.7532} \\
 & \shade AICL* & \shade 3.35 & \shade .8496 & \shade .9334 & \shade .8616 & \shade 1.27 & \shade .9337 & \shade .9294 & \shade .9313 & \shade 2.25 & \shade .9513 & \shade .9367 & \shade .9413 \\ 	
\cmidrule{2-14}
\multirow{5}{*}{\rot{CoLA}} & 0-shot & 0 & .5585 & .5323 & .3699 & 0 & .4315 & .4899 & .2558 & 0 & .6474 & .4321 & .3455 \\ 		 
 & FICL & 10 & .6240 & .6415 & .6008 & 9 & .7071 & .6306 & .6433 & 8 & .7167 & .6338 & .6472 \\ 
 & AICL(E) & 8.32 & .6556 & .6741 & .6582 & 7.73 & .7289 & .6471 & .6601 & 2.33 & .7412 & .6447 & .6667 \\ 
 & AICL(E+N) & 7.43 & \textbf{.6580} & \textbf{.6765} & \textbf{.6616} & 7.93 & \textbf{.7392} & \textbf{.6486} & \textbf{.6613} & 1.99 & \textbf{.7432} & \textbf{.6558} & \textbf{.6714} \\ 
 & \shade AICL* & \shade 2.42 & \shade .9417 & \shade .9281 & \shade .9299 & \shade .69 & \shade .9663 & \shade .9232 & \shade .9413 & \shade .78 & \shade .9408 & \shade .9466 & \shade .9483 \\ 				
\cmidrule{2-14}
\multirow{5}{*}{\rot{RTE}} & 0-shot & 0 & .4913 & .4992 & .3985 & 0 & .6741 & .6741 & .6741 & 0 & .5345 & .5444 & .4403 \\ 		
 & FICL & 10 & .6095 & .6051 & .5967 & 1 & .7347 & .7239 & .7240 & 4 & .6287 & .6233 & .6214 \\ 
 & AICL(E) & 8.57 & .6304 & .6277 & .6227 & 2.44 & .7536 & .7412 & .7415 & 5.16 & .7592 & .6441 & .6700 \\ 	
 & AICL(E+N) & 8.32 & \textbf{.6342} & \textbf{.6311} & \textbf{.6252} & 1.15 & \textbf{.7551} & \textbf{.7465} & \textbf{.7471} & 3.68 & \textbf{.7631} & \textbf{.6485} & \textbf{.6738} \\
 & \shade AICL* & \shade .95 & \shade .9783 & \shade .9783 & \shade .9783 & \shade .81 & \shade .9296 & \shade .9214 & \shade .9234 & \shade .65 & \shade .9734 & \shade .9733 & \shade .9733 \\ 	
\bottomrule
\end{tabular}
\end{adjustbox}
\caption{
\small
A comparison between the two variants of our proposed Adaptive ICL approach and the baselines on the 4 datasets used in our experiments.
The column `$k$' denotes the number of few-shot examples. In particular, for AICL, this denotes the average of the number of examples used for each test instance. In our experiments, the maximum value of $k$ (i.e., $M$ of Equation \ref{eq:candidate}) is set to 10 for both AICL and FICL.
We found that the AICL(E+N) results are statistically distinguishable with respect to the FICL results (McNemar’s test with $p=95\%$ confidence interval).
\label{table:results}
}
\end{table}

%% file: figdefs/results.tex
\begin{figure}[t]
    \centering
    \begin{subfigure}[t]{0.24\textwidth}
        \centering
        \includegraphics[width=.99\linewidth]{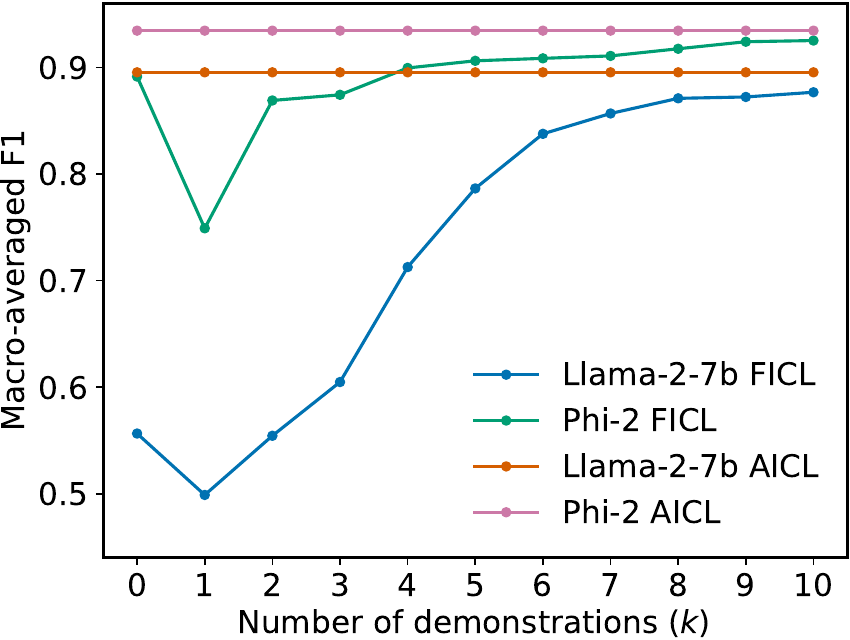}
        \caption{SST2}
    \end{subfigure}
    \begin{subfigure}[t]{0.24\textwidth}
        \centering
        \includegraphics[width=.99\linewidth]{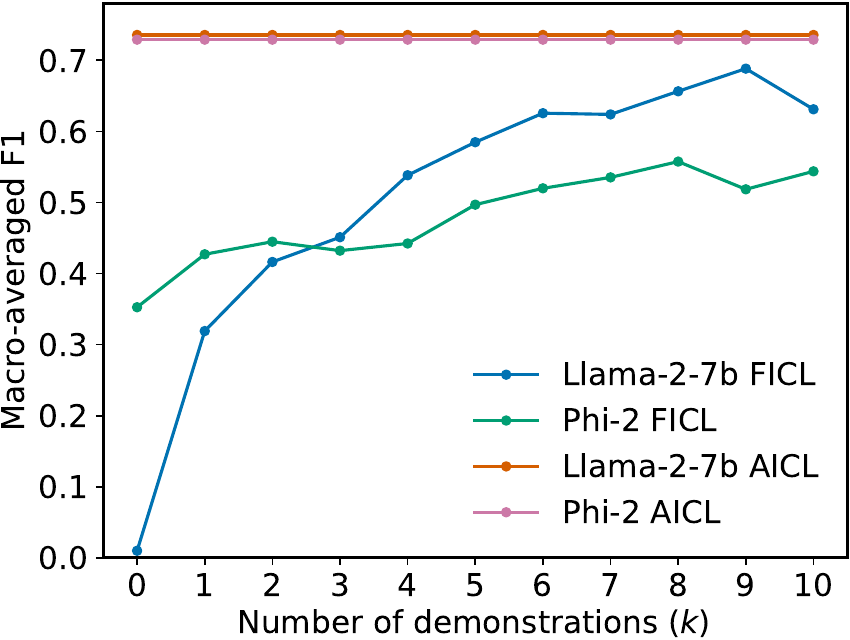}
        \caption{TREC}
    \end{subfigure}
    \begin{subfigure}[t]{0.24\textwidth}
        \centering
        \includegraphics[width=.99\linewidth]{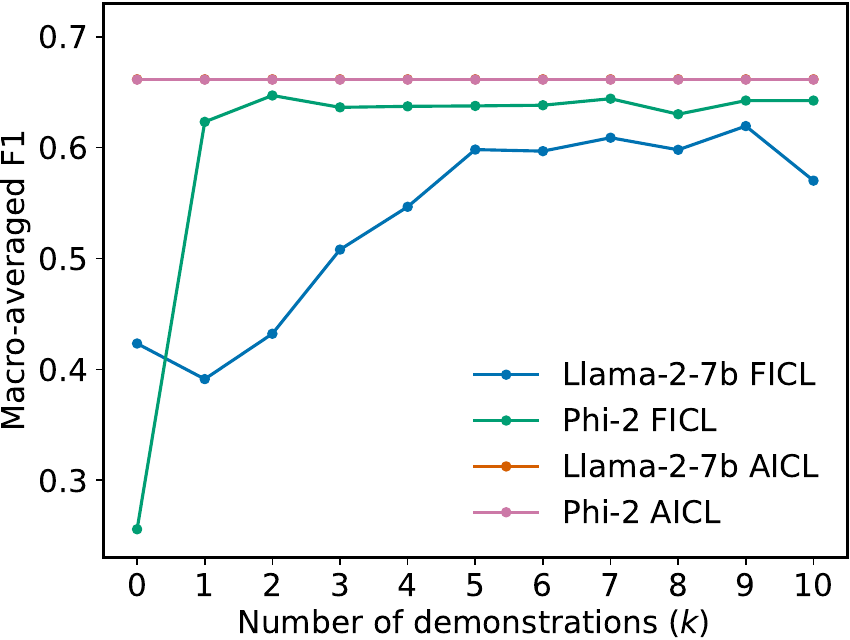}
        \caption{CoLA}
    \end{subfigure}
    \begin{subfigure}[t]{0.24\textwidth}
        \centering
        \includegraphics[width=.99\linewidth]{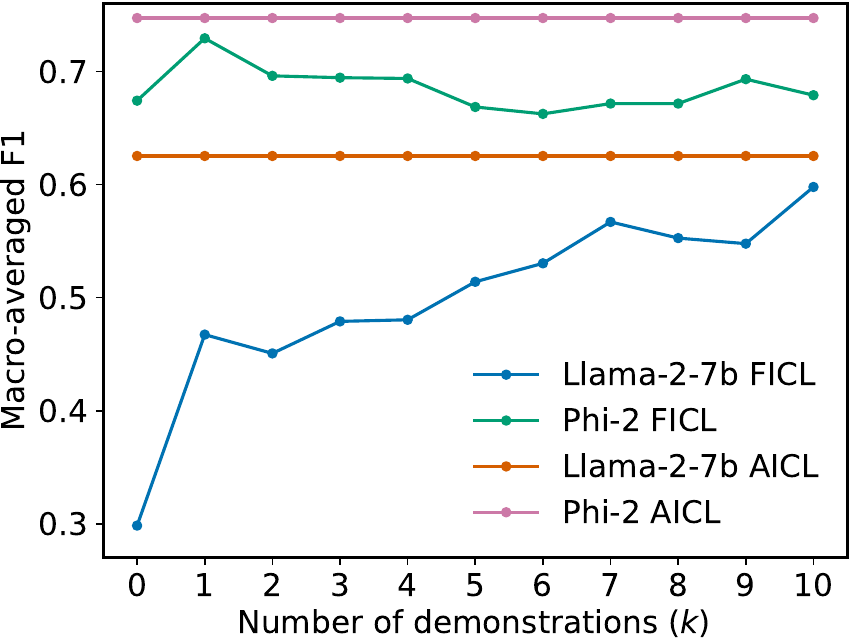}
        \caption{RTE}
    \end{subfigure}
    \caption{    
    FICL (macro-averaged) F-scores for different context sizes on the test splits of the different datasets (AICL results also included for comparison).
    These plots demonstrate that AICL can be applied on any dataset without requiring to optimize any hyper-parameter (e.g., $k$ in FICL).
    }
    \label{fig:results}
\end{figure}

%% file: figdefs/scalability.tex
\begin{figure}[t]
    \centering
    \begin{subfigure}[t]{0.24\textwidth}
        \centering
        \includegraphics[width=.99\linewidth]{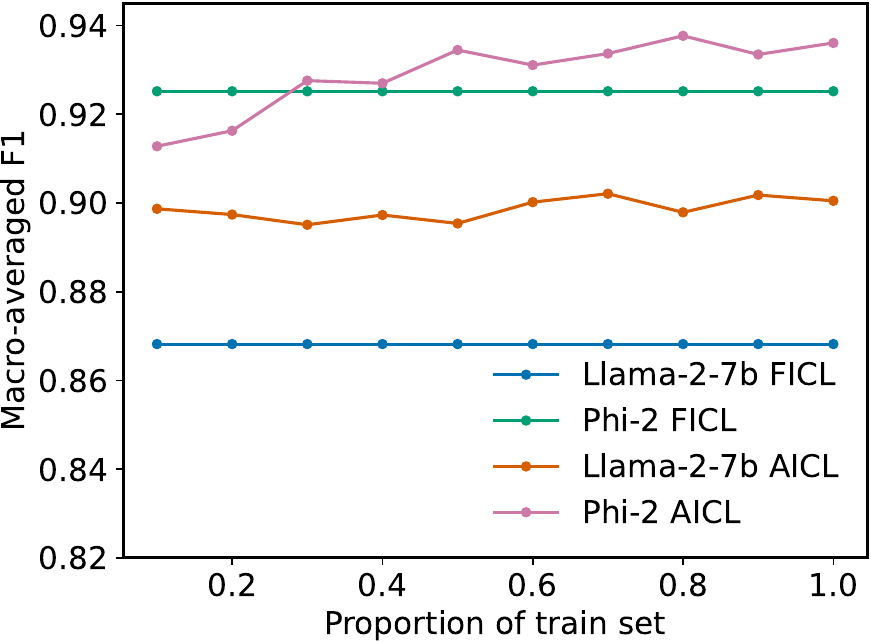}
        \caption{SST2}
    \end{subfigure}
    \begin{subfigure}[t]{0.24\textwidth}
        \centering
        \includegraphics[width=.99\linewidth]{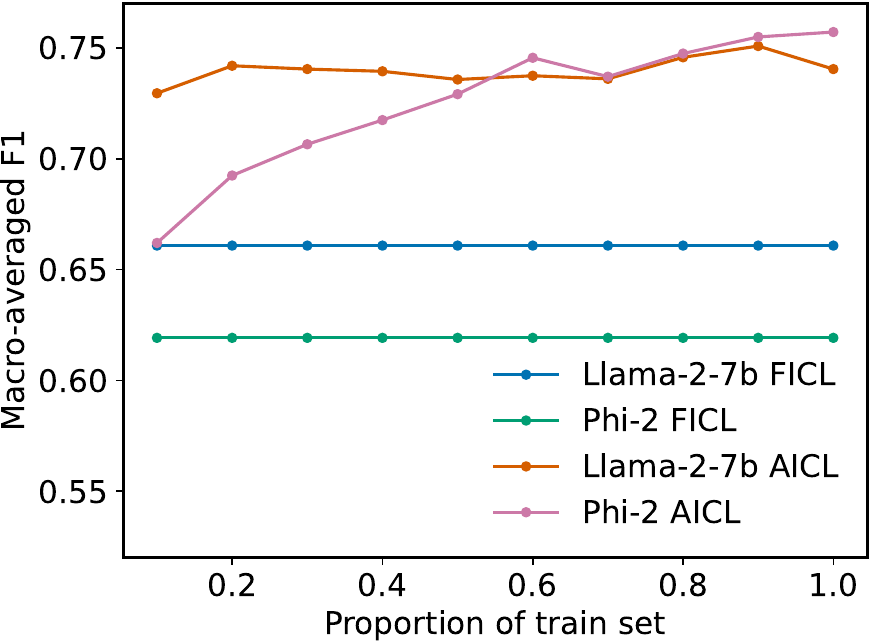}
        \caption{TREC}
    \end{subfigure}
    \begin{subfigure}[t]{0.24\textwidth}
        \centering
        \includegraphics[width=.99\linewidth]{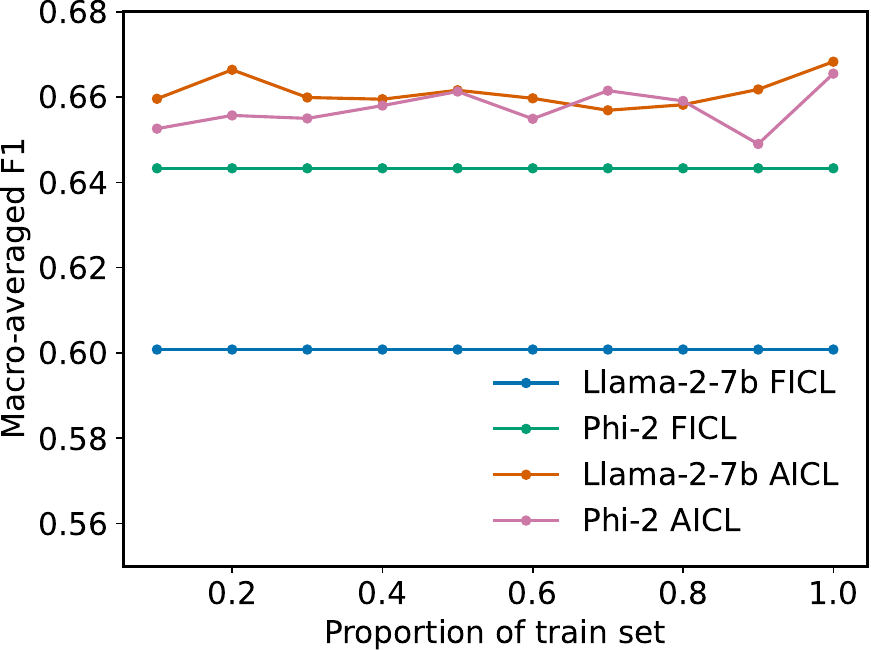}
        \caption{CoLA}
    \end{subfigure}
    \begin{subfigure}[t]{0.24\textwidth}
        \centering
        \includegraphics[width=.99\linewidth]{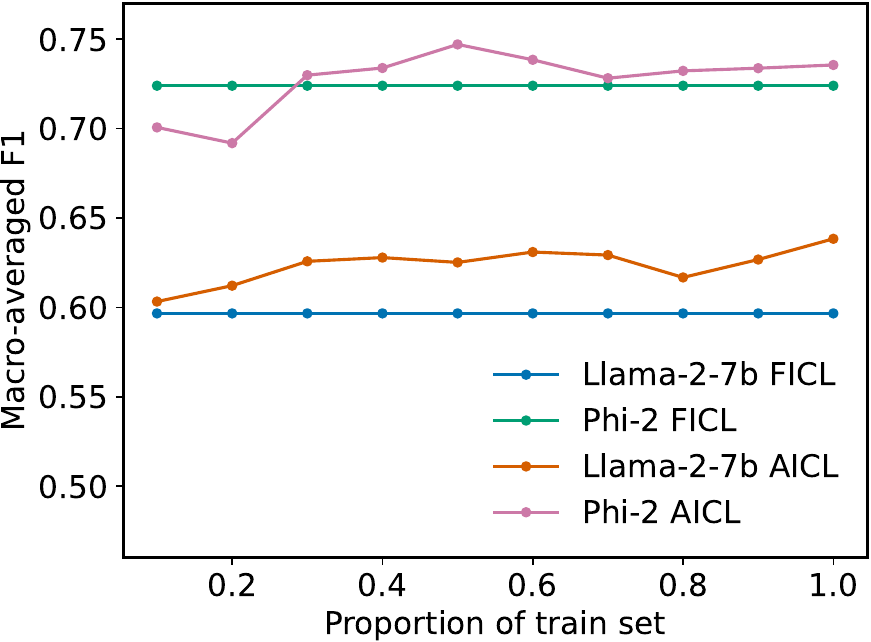}
        \caption{RTE}
    \end{subfigure}
    \caption{
    Macro-averaged F1 scores with different proportions of the training set used to train $\kappa(\vect{x})$ - the multi-label classifier of AICL (Equation \ref{eq:mlc-l}).
    }
    \label{fig:scalability}
\end{figure}

%% file: tabledefs/results_nolabel.tex
\begin{table}[t]
\centering
\begin{adjustbox}{width=0.72\textwidth}
\small
\begin{tabular}{@{}l@{~~}l@{~~}l@{~~}l@{~~}l@{~~}l@{~~}l@{~~}l@{}}
\toprule
 & & \multicolumn{3}{c}{RAG setup (w/o Labels)} & \multicolumn{3}{c}{ICL setup (w/ Labels)} \\ \cmidrule(r){3-5} \cmidrule(r){6-8}
Dataset & 0-shot & FICL & AICL(E) & \shade AICL* & FICL & AICL(E) & \shade AICL* \\
\midrule
SST2 & .8914 & .7339 & \textbf{.9119} & \shade .9610 & .9252 & .9300 & \shade .9863  \\
TREC & .3526 & .4287 & \textbf{.4752} & \shade .4922 & .6192 & .7196 & \shade .9313 \\
CoLA & .2558 & .2469 & \textbf{.2679} & \shade .7937 & .6433 & .6601 & \shade .9413 \\
RTE & \textbf{.6741} & .6144 & .6688 & \shade .8655 & .7240 & .7415 & \shade .9234 \\

\bottomrule
\end{tabular}
\end{adjustbox}
\caption{
A comparison between the F1 scores obtained by AICL (Phi-2) on two different settings - a) w/ labels (same as Table \ref{table:results}), and b) w/o labels, as in a RAG pipeline.
Best performing RAG results are bold-faced.
\label{table:results_nolabel}
}
\end{table}

%% file: sections/conclusion.tex
To improve the effectiveness of LLM-based few-shot inference, we proposed a supervised multi-label classification approach that learns how many examples to use for which data instance. The classifier itself is trained on a set of labeled examples, where the ground-truth for each training data instance constitutes the set of $k$ (number of examples) that lead to correct downstream task prediction via $k$-shot LLM inference.
%
